%% file: main.tex
\documentclass[twoside]{article}

\usepackage{aistats2020}

\input{math_commands.tex}

\usepackage{hyperref}
\usepackage{url}
\usepackage[ruled]{algorithm2e}
\usepackage{svg}
\usepackage{tabularx}
\usepackage{placeins}

\usepackage[round]{natbib}

\begin{document}

\twocolumn[

\aistatstitle{Computationally Efficient Feature Significance and Importance for Machine Learning Models}



\aistatsauthor{ Enguerrand Horel \And Kay Giesecke}

\aistatsaddress{ Institute for \\ Computational and Mathematical Engineering \\ Stanford University \And Department of \\ Management Science and Engineering \\ Stanford University } 
]

\begin{abstract}
We develop a simple and computationally efficient significance test for the features of a machine learning model. Our forward-selection approach applies to any model specification, learning task and variable type. The test is non-asymptotic, straightforward to implement, and does not require model refitting. It identifies the statistically significant features as well as feature interactions of any order in a hierarchical manner, and generates a model-free notion of feature importance. Experimental and empirical results illustrate its performance.
\end{abstract}

\section{Introduction}
\label{S1}

Machine learning models are notoriously hard to interpret. This is an important issue in many application domains. Superior predictive performance alone is not sufficient; model stakeholders require interpretable models to make model-driven decisions. In this paper, we develop a computationally efficient method to assess the statistical significance and the importance of the features of a machine learning model. The method provides statistical insight into a ``black-box" model.


\textbf{Main contributions.} Our significance test is based on a novel application of a forward-selection approach. It can be directly applied to a trained model, does not require model refitting, and is easily implemented. The test compares the loss for a fitted model when only an intercept is active with the loss when both an intercept as well as a feature are active. The loss difference captures the intrinsic contribution of the feature in isolation, leading to an informative notion of feature importance. We call this test procedure Single Feature Introduction Test (SFIT). This approach has the advantage of being robust to correlation among features. Additionally, we introduce an auxiliary type-I error control parameter which prevents false discoveries due to a model's susceptibility to non-informative features. Other advantages of our method include: (1) it does not make any assumptions regarding the distribution of the data nor the specification of the model; (2) it can be applied to both regression and classification problems and both numeric and categorical features; (3) it naturally extends to the analysis of higher order feature interactions in a hierarchical manner. For the specific case of a neural network, we propose a strategy to reduce the search space for testing significant higher-order interactions by leveraging links among the network parameters.

\textbf{Related literature.} Most previous work that seeks to explain model predictions assesses the relative contribution of features. Examples include \cite{kononenko2010efficient},  \cite{baehrens2010explain}, \cite{cortez2011opening}, \cite{ribeiro2016should}, \cite{datta2016algorithmic}, \cite{lundberg2017unified}.\footnote{There are a number of model-specific approaches. For neural networks see, for example,  \cite{shrikumar2017learning},
 \cite{olden2002illuminating},   \cite{sundararajan2017axiomatic},   \cite{horel2018sensitivity}, \cite{williamson2017nonparametric}.} \cite{guidotti2018survey} provides a review. Our work also offers a measure of feature importance, but additionally provides a rigorous statistical test of feature significance. 
 
 Existing tests of feature significance tend to be computationally expensive or are only valid asymptotically. For instance, a likelihood ratio test can be used to perform asymptotic inference (\cite{vuong1989likelihood}). This approach is computationally demanding because it requires fitting a model for each individual variable to be tested. Moreover, in case of model mis-specification, the asymptotic distribution is challenging to compute, especially in the high-dimensional settings that are the norm rather than the exception. Other general non-parametric testing methods include goodness-of-fit tests, conditional moment tests (see Section 6.3 of \cite{henderson2015applied} for a review) or derivative-based tests such as  \cite{racine1997consistent}, \cite{horel2019towards}. To perform these tests, the null distribution of the test statistic is either estimated via bootstrap or is derived asymptotically under a specific choice of non-parametric model. In contrast, our method provides statistical guarantees that hold in finite-sample regimes without depending on a specific model or requiring a bootstrap procedure. Furthermore, because our method does not require model refitting, it is attractive for models that require a relatively long training time such as neural networks. 
 
 An alternative finite-sample approach has been developed in the framework of ``knockoffs," see  \cite{barber2015controlling}, \cite{candes2018panning}, and \cite{barber2018robust}. This feature selection procedure provides false discovery rate control and is partially model-free in the sense that it does not make any assumptions regarding the distribution of the outcome conditional on the inputs. However, it requires to either specify the distribution of the features or to approximate it. This approach requires including additional ``knockoff features" during training which can alter the predictive performance of the model. Similar approaches proposed by  \cite{tansey2018holdout} and \cite{burns2019interpreting} also require the ability to sample from the conditional distribution of the feature to be tested given the remaining variables. In contrast, our method directly applies to any fitted model, does not impose assumptions on the distribution of the features, and does not require the ability to sample from a specific distribution. 

Our approach is inspired by the Leave-One-Covariate-Out (LOCO) method of \cite{Lei2018predictive}. These authors propose an interesting finite-sample, model-free inference framework. Their approach entails removing the feature of interest from the feature set and evaluating the impact on predictive performance. Unfortunately, this approach suffers from the same computational inefficiency as a likelihood ratio test. Moreover, this approach is highly susceptible to collinearity. This is an important issue because one often has high-dimensional feature sets with correlated and redundant features. To illustrate the issue, consider the following toy regression: $Y = 2 + X_1 + X_2 + \epsilon$, where $(X_1,X_2) \sim \mathcal{N}((0,0)^\top,\Sigma)$ and $\Sigma$ is a covariance matrix with off-diagonal elements equal to 0.85 and $\epsilon \sim \mathcal{N}(0,10)$. Since the variables are highly correlated, we have $Y \approx 2 + 2X_1 + \epsilon \approx 2 + 2X_2 + \epsilon$, meaning that removing a variable is unlikely to significantly impact  predictive performance. Simulation tests show that the LOCO method considers each variable as insignificant half of the time (0.503 and 0.498).\footnote{Based on averaging the results of 3,000 tests, each time performed using new simulated training and inference sets of 1,000 samples each.}  
The SFIT method we propose identifies both variables as significant for all simulations, meaning perfect power.   

Section \ref{S2} develops SFIT and its implementation. Section \ref{S3} illustrates its performance through simulations. Section \ref{S4} provides an empirical application to a credit default dataset.

\section{Single Feature Introduction Test (SFIT) procedure}
\label{S2}

\subsection{Description}
\label{S21}

Consider i.i.d. data $Z_1,...,Z_n \sim P$ with $Z_i = (X_i, Y_i)$. $X_i$ is a random vector of size $(p+1)$ that contains the $p$ features plus an intercept at the first coordinate. The features can be a mix of numeric and categorical variables, the latter ones being one-hot encoded. $Y_i$ can be either a real number in the case of a regression or an index from $\{1,2,...,C\}$ in a classification setting. 
We split $\{1,...,n\}$ into two subsets $\mathcal{I}_1$ and $\mathcal{I}_2$ and let $\mathcal{D}_k = \{(X_i,Y_i): i \in \mathcal{I}_k\}$, $k = 1,2$.

We are given a predictive model $\mu$ that has been trained on $\mathcal{D}_1$. This model will typically estimate the conditional mean $ \mathbb{E}(Y|X)$ in a regression or the conditional probabilities $\mathbb{P}(Y = \cdot | X)$ in a classification. To evaluate the importance of an individual feature $j$, we define the transformed vector $X_i^{(1,j)}$, which is obtained from $X_i$ by setting all entries but the $j^{th}$ and the intercept to $0$. We abbreviate $X_i^{(1)}=X_i^{(1,0)}$; this is the vector for which all entries but the intercept are 0. 
We choose a loss function $L(Y,\mu(X))$. For example, for a regression we may take the absolute deviation $L(Y,\mu(X)) = |Y - \mu(X)|$ while in a classification the cross entropy $L(Y,\mu(X)) = -\log(\{\mu(X)\}_{Y})$. 

For $\beta\in[0,1]$, consider 
\begin{multline*}
\Delta_j^i = \Delta_j(X_i,Y_i) \\ 
= (1-\beta)L\Big(Y_{i},\mu\big(X_i^{(1)}\big)\Big) - L\Big(Y_{i},\mu\big(X_i^{(1,j)}\big)\Big).
\end{multline*}
This is the difference between $(1-\beta)\%$ of the loss using the intercept as the only predictor (the baseline) and the loss using the intercept plus the feature $j$. We consider $(1-\beta)$ times the baseline loss rather than the loss itself to make the test more robust to non-informative variables and control its type-I error. Below we will further justify the role of $\beta$. 
Further define the statistic $\hat{m}_j$, which will be used to assess the significance of feature $j$:
$$\hat{m}_j = \text{median}_{i \in \mathcal{D}_2} \Delta_j^i $$ for $2 \leq j \leq p + 1$.
This is the median over the inference set $\mathcal{D}_2$ of the loss differences. Intuitively, $\hat{m}_j$ represents the predictive power of variable $j$ compared to a baseline model that only has an intercept term. A larger value is associated with higher predictive power. 

There are two fundamental differences between the SFIT statistic and other related significance statistics such as the likelihood ratio or the LOCO statistic. (1) Rather than removing one feature at a time to assess significance, we introduce one feature at a time to evaluate its impact relative to the baseline intercept. This approach has two advantages. First, it defines a more intuitive notion of variable importance by measuring the exclusive effect of a feature on the prediction. 
Another advantage is robustness to correlation among features. If two variables are highly correlated but important, our method will identify both as significant. (2) By applying the fitted model on the masked vector $X_i^{(1,j)}$, we eliminate the need to refit a new model on the intercept and the $j^{th}$ variable. The obvious advantage of this is computational efficiency, which is rather important for models that can take a long time to train (e.g. neural networks). It further allows us to perform inference directly on the model of interest and not on a newly trained one. Finally, by zeroing out all the other variables, we are guaranteed to capture the exclusive effect of a variable.

The choice of setting variables to the value 0 is somewhat arbitrary; what matters most is to constraint the variables to keep the same value across all variables. Indeed, what is being tested in this procedure is whether or not the variance from different values of a variable adds predictive power over keeping a constant value. Because it is common practice to center and scale continuous variables in pre-processing, zeroing them out has the interpretation of setting variables to their mean value. However, for one-hot encoded categorical variables, this approach can understate significance in cases where most values are already equal to 0. In those cases, we recommend setting the values equal to a number strictly greater than zero and strictly smaller than one rather than zero.

\subsection{Statistical inference}
\label{S22}

The statistic $\hat{m}_j$ will be used to construct a finite-sample significance test. We assume that the distribution of $\Delta_j(X,Y)$ conditional on the training set $\mathcal{D}_1$ is continuous and consider its median, 
$$m_j = \text{median}_{(X,Y)}\big[ \Delta_j(X,Y) | \mathcal{D}_1\big].$$
Our goal is to obtain finite-sample confidence interval for $m_j$ and perform the following one-sided hypothesis test of significance:
$$H_0 : m_j \leq 0 \; \text{versus} \; H_1: m_j > 0$$ using the statistic $\hat{m}_j$. Following the procedure of LOCO \cite{Lei2018predictive}, we perform a standard sign test (\cite{dixon1946sign}) by counting the number of times $n_j^{+}$ that $\Delta_j^i > 0$ for $i \in \mathcal{I}_2$. Under the null, $n_j^{+}$ follows a binomial distribution with parameters $n_2$ and $1/2$ where $n_2=|\mathcal{D}_2|$. Hence this null can be tested using a one-sided (greater) binomial test. Confidence intervals for $m_j$ can be obtained by inverting the sign test and using the order statistics $\Delta_j^{(i)}$. This leads to a sequence of nested intervals $[\Delta_j^{(1)},\Delta_j^{(n_2)}]$, $[\Delta_j^{(2)},\Delta_j^{(n_2-1)}]$ where the $k^{th}$ confidence interval $[\Delta_j^{(1+k)},\Delta_j^{(n_2-k)}]$ will have an exact coverage equal to $1 - 2\mathbb{P}(B \leq k)$, where $B \sim \mathcal{B}(n_2,1/2)$. These confidence intervals have exact finite sample coverage but the coverage level $\alpha$ cannot be chosen exactly but has to be chosen among the values $\alpha_k = 2\mathbb{P}(B \leq k)$. An alternative is to consider asymptotically valid confidence intervals with coverage approximately equal to $1-\alpha$ (due to ceiling and flooring) given by $[\Delta_j^{\lfloor \frac{n_2+1}{2}-q_{1-\alpha/2}\frac{\sqrt{n_2}}{2}\rfloor},\Delta_j^{\lceil \frac{n_2+1}{2}+q_{1-\alpha/2}\frac{\sqrt{n_2}}{2}\rceil}]$ where $q_{\alpha}$ is the $\alpha$-quantile of a standard normal variable.

Algorithm \ref{algo 1} provides the steps required to test  significance and obtain confidence intervals.

\begin{algorithm}
\footnotesize
\SetAlgoLined
\KwIn{Model $\mu$, dataset $\mathcal{D}_2 = \{X_i,Y_i\}_{i \in \mathcal{I}_2}$, significance level $\alpha$, $\beta$, function $\textsc{BinomTest}$($\#$successes, $\#$trials, hypothesized probability of success, alternative hypothesis) that outputs the p-value of a binomial test}
\KwOut{Set of first order significant variables $\mathcal{S}_1$ and their related confidence intervals $\mathcal{C}_1$}
$\mathcal{S}_1 = \emptyset$, $\mathcal{C}_1 = \emptyset$, generate the masked dataset $\{X_i^{(1)}: i \in \mathcal{I}_2\}$ \;
 \For{$j = 2$  \KwTo  $p+1$}{
    Generate the masked dataset $\{X_i^{(1,j)}: i \in \mathcal{I}_2\}$ \;
    Compute $\Delta_j^i = (1-\beta)L\Big(Y_{i},\mu\big(X_i^{(1)}\big)\Big) - L\Big(Y_{i},\mu\big(X_i^{(1,j)}\big)\Big)$ for all $i \in \mathcal{I}_2$ \;
    $n_j^{+} = \sum_{i\in \mathcal{I}_2} 1_{\{\Delta_j^i > 0\}}$ \;
    \If{$\textsc{BinomTest}(n_j^{+},n_2,1/2,\text{greater}) < \alpha$}{
        $\mathcal{S}_1 = \mathcal{S}_1 \cup \{j\}$ \;
        $\mathcal{C}_1[j] = [\Delta_j^{\lfloor \frac{n_2+1}{2}-q_{1-\alpha/2}\frac{\sqrt{n_2}}{2}\rfloor},\Delta_j^{\lceil \frac{n_2+1}{2}+q_{1-\alpha/2}\frac{\sqrt{n_2}}{2}\rceil}]$ \;
    }
 }
 \caption{First-order SFIT}
 \label{algo 1}
\end{algorithm}


The only parameters one has to choose to perform the test is the significance level $\alpha$, the parameter $\beta$ and eventually the relative sizes of the training set $\mathcal{D}_1$ and test set $\mathcal{D}_2$ if the model has not been trained yet. The significance level $\alpha$ can be set at a fixed value chosen by the user such as 0.05 or 0.01 or can be determined adaptively to control false discoveries of significant features. Indeed, in a high-dimensional setting where many tests would be performed, false discoveries are more likely. Because of the dependency of our tests, we suggest using the FDR control method under dependency described in \cite{benjamini2001control}. The size of the test set $\mathcal{D}_2$ should be at least several thousand to be able to detect features or interactions that have small predictive effect as demonstrated in \cite{dixon1946sign}.

The parameter $\beta$ is crucial for testing: it makes the test more robust to false discoveries due to potential model over-fitting. For instance, recent works \cite{zhang2016understanding}, \cite{wei2018margin} have shown that despite having good generalization properties, deep neural networks often tend to be over-parametrized and can overfit the training set. In practice, over-parametrization implies that the model might over-utilize non-informative features. Increasing the degree of model regularization can mitigate this issue. However, since we consider the model as given, we cannot control the regularization during training. Therefore, instead of regularizing the model, we regularize the test through the $\beta$ parameter. Increasing $\beta$ will reduce the number of non-informative features selected. However, after a certain point, the test will fail to select the informative variables by being too stringent. Hence, $\beta$ can control the trade-off between false discoveries and false rejections. Our numerical results indicate that without $\beta$, the test would suffer from high type-I errors, i.e. considering many non-informative features as significant. A user can decide on a (typically small) value for $\beta$ by determining the tolerance for false discoveries vs. false rejections. 

 We provide a data-driven approach to choosing $\beta$ which is inspired by the parameter randomization test described in \cite{adebayo2018sanity}. The idea is based on the following observation: a model whose parameter values have been chosen randomly rather than through a fitting procedure should not have any significant predictive power. Any predictive effect exhibited by the random model should be considered as noise. $\beta $ can be chosen to ensure that our method applied to a random model would consider all features as non-significant. More specifically, we can loop over increasing values of $\beta$ on a logarithmic scale. For each value, we generate many models with the same structure as the model to be tested but with randomly chosen parameter values. We compute the average false discovery rate (ratio of features found as significant by SFIT over the random models) and compare it with $\alpha$. If it is smaller than $\alpha$, then $\beta$ is large enough and we can stop. Otherwise, the next largest value of $\beta$ is chosen and the procedure is repeated. The test set $\mathcal{D}_2$ should be split into two subsets: a validation set $\mathcal{D}_v$ used to determine $\beta$ and a test set $\mathcal{D}_2$. Given an optimal $\hat{\beta}$, the tests can now be performed on the test set $\mathcal{D}_2$. The only difference to what was presented before is that inference is now done on the median of the distribution of $\Delta_j$ conditioned on both $\mathcal{D}_1$ and $\mathcal{D}_v$.

\subsection{Higher order effects}
\label{S23}

The SFIT method allows one to obtain a hierarchical representation of feature importance. Indeed, Algorithm \ref{algo 1} described in the previous section selects the variables that are significant by themselves but can fail to select the variables that interact with other variables. As an illustrative example, let's consider the following model $\mu(x_1,x_2,x_3) = 1 + x_1x_2+ 2x_3$. Only the third feature would be considered as significant here because setting one of the first two inputs to zero annihilates the effect of the other. This means that a variable considered as non-significant by a first-order SFIT method might still have predictive power through a second or higher order interaction.

Given the set of features selected by a first-order SFIT, we can first check that there are some significant higher-order interactions that potentially need to be uncovered. Similar to the previous procedure, one can look at the prediction error difference between two models: (1) the model that includes only first-order significant variables and (2) the model that includes all variables, and test whether this difference is significantly greater than zero.

If this test highlights the predictive power of higher-order effects, then we can start looking for variables that would impact the outcome through second-order interactions. To do so, we would go over all the variables considered as non significant by the first-order SFIT, pair them with every other variable one at a time and test for the significance of the resulting interaction. A distinction has to made whether the variable to be tested is paired with a first order significant variable selected by Algorithm \ref{algo 1} or with an other non-significant variable. In the former case, if we denote by $j$ the variable to be tested and by $k$ the significant variable it is paired with, then the metric $\Delta^i_j$ has to be updated as $\Delta_j^i(X_i,Y_i) = (1-\beta)L\Big(Y_i,\mu\big(X_i^{(1,k)}\big)\Big) - L\Big(Y_i,\mu\big(X_i^{(1,j,k)}\big)\Big)$.
In the latter case where variable $j$ is paired with $k$ that is also non-significant, the metric $\Delta_j$ is simply $\Delta_j^i(X_i,Y_i) = (1-\beta)L\Big(Y_i,\mu\big(X_i^{(1)}\big)\Big) - L\Big(Y_i,\mu\big(X_i^{(1,j,k)}\big)\Big)$. The implementation of this procedure is described in Algorithm \ref{algo 2}.

\begin{algorithm}
\footnotesize
\SetAlgoLined
\KwIn{Model $\mu$, dataset $\mathcal{D}_2 = \{X_i,Y_i\}_{i \in \mathcal{I}_2}$, significance level $\alpha$, $\beta$, function $\textsc{BinomTest}$ that outputs the p-value of a binomial test, set of significant inputs $\mathcal{S}_1$ obtained by first-order SFIT, set of ordered expected interactions for each feature $j$: $\{\mathcal{P}_j\}_{j \in \{1,...,p\}}$}
\KwOut{Set of second-order significant variables $\mathcal{S}_2$ along with their associated pairs and confidence intervals $\mathcal{C}_2$}
$\mathcal{S}_2 = \emptyset$, $\mathcal{C}_2 = \emptyset$, $\mathcal{U}_1 = \{1,...,p\}\setminus \mathcal{S}_1$ \;
\emph{First check that there are some significant higher-order effects}\;
Compute $\Delta_{\mathcal{U}_1}^i = L\Big(Y_{i},\mu\big(X_i^{(\mathcal{S}_1)}\big)\Big) - L\Big(Y_{i},\mu\big(X_i\big)\Big)$ for all $i \in \mathcal{I}_2$ \;
    $n_j^{+} = \sum_{i\in \mathcal{I}_2} 1_{\{\Delta_{\mathcal{U}_1}^i > 0\}}$ \;
\If{$\textsc{BinomTest}(n_j^{+},n_2,1/2,\text{greater}) > \alpha$}{
   \Return{$\mathcal{S}_2$, $\mathcal{C}_2$}
}
 \For{$j \in \mathcal{U}_1$ and $j \notin \mathcal{S}_2$}{
    \For{$k \in \mathcal{P}_j$}{
    \eIf{$k \in \mathcal{S}_1$}{
        Compute $\Delta_{(j,k)}^i = (1-\beta)L\Big(Y_{i},\mu\big(X_i^{(1,k)}\big)\Big) - L\Big(Y_{i},\mu\big(X_i^{(1,j,k)}\big)\Big)$ for all $i \in \mathcal{I}_2$ \;
    }{
        Compute $\Delta_{(j,k)}^i = (1-\beta)L\Big(Y_{i},\mu\big(X_i^{(1)}\big)\Big) - L\Big(Y_{i},\mu\big(X_i^{(1,j,k)}\big)\Big)$ for all $i \in \mathcal{I}_2$ \;
    }

    $n_{(j,k)}^{+} = \sum_{i\in \mathcal{I}_2} 1_{\{\Delta_{(j,k)}^i > 0\}}$ \;
    \If{$\textsc{BinomTest}(n_{(j,k)}^{+},n_2,1/2,\text{greater}) < \alpha$}{
        $\mathcal{S}_2 = \mathcal{S}_2 \cup \{j\}$ \;
        \lIf{$k \in \mathcal{U}_1$}{$\mathcal{S}_2 = \mathcal{S}_2 \cup \{k\}$}
        $\mathcal{C}_2[(j,k)] = [\Delta_{(j,k)}^{\lfloor \frac{n_2+1}{2}-q_{1-\alpha/2}\frac{\sqrt{n_2}}{2}\rfloor},\Delta_{(j,k)}^{\lceil \frac{n_2+1}{2}+q_{1-\alpha/2}\frac{\sqrt{n_2}}{2}\rceil}]$ \;
        }
    }
 }
 \caption{Second-order SFIT}
 \label{algo 2}
\end{algorithm}

This higher-order procedure is designed to reveal the variables that have a second or higher-order significance through interactions with other variables but that are not first-order significant. Overall, the goal of this method is to discover all the significant variables and their orders of significance which is distinct from discovering all significant interactions. This is why we do not exhaustively test for all possible interactions between variables. If one variable has been tested as first-order significant, we consider that any potential higher-order significance that this variable could have in addition, do not matter since this variable is already highlighted as significant. However, if one's goal is to discover all the significant interactions between variables, the higher-order procedure previously described can be easily adapted by going over all the variables and not only over the ones considered as non significant by the first-order SFIT.

Since we expect the set of non-significant variables obtained from the first-order SFIT to be smaller than $p$, the number of pairs to consider for second-order SFIT should stay smaller than $p^2$. However the worst case scenario could still be of the order $p^2$, which can be computationally prohibitive in high-dimensional settings. To overcome this issue, we propose the following speed-up strategy for neural networks specifically. By looking at the parameters matrix $W_{h,p+1}$ corresponding to the linear map that goes from the input layer of size $(p+1)$ to the first hidden layer of size $h$, we can flag the pairs of variables that are likely to interact with each other. Indeed, if the weights of the edges coming from feature $j$ and feature $k$ to hidden node $h$ have large values, then it is likely that inputs $j$ and $k$ will interact. A quantitative way to measure this potential interaction is by computing the empirical covariance of the absolute parameters matrix $|W|$, $|W|^T|W|$ where $|\cdot|$ represents the entry-wise absolute value operator. Computing this covariance matrix still takes $hp^2$ operations, but in most of programming languages this operation can be done using BLAS functions that would take only a couple of seconds even for $p$ and $h$ of the order of thousands in a regular laptop. $|W|^T|W|_{j,k}$ can be used to measure the potential of interaction between features $j$ and $k$. Then, for each feature $j$, we can look at the sorted $j^{th}$ row of $|W|^T|W|$ and restrict the pairs to consider to the $l^{th}$ largest values. $l$ would be a parameter chosen by the user as a function of the dimension of the problem and the computational power available. In a common case where $l << p$, finding all the second-order significant features is done in a computing time linear in the number of features.

Overall, the SFIT method give us a hierarchical ranking of the importance of the features. The first-order SFIT identifies all the variables that have a first order effect on the outcome. Then the second-order SFIT selects the variables having a significant second-order interaction with another variable but that does not have a first-order effect. At each step, a global significance test of the currently identified features can be done to decide whether or not a higher-order SFIT should be run on the model. The method was only described until the second-order in Algorithms \ref{algo 1} and \ref{algo 2}, but can be naturally extended to higher-order interactions.

\subsection{Extensions}
\label{S24}
Our method can be extended to different settings. 

One way is to partition the inference set $\mathcal{D}_2$ to obtain finer level of tests. For instance, in the case of a multi-class classification problem, one can obtain the significant variables per class by running the SFIT method on the observations from $\mathcal{D}_2$ that belongs to each class. Another use-case is when the dataset consists of a population with different subgroups. One might be interested in analysing each subgroups independently. Partitioning $\mathcal{D}_2$ according to the subgroups and applying our procedure on each partition inform of the important variables per subgroup.

Another way to extend this method is to consider different choice of baseline losses to compute the $\Delta^i_j$. We presented the case where the baseline is defined by the intercept term only. Using such a baseline informs on the predictive value of adding one variable compared to not having any variables. However, in some cases it might be interesting to assess the impact of adding one new variable to a given set of variables. This can be done by computing the baseline loss on the data where all the variables are set to zero except for the ones in the given set.  

\section{Simulation results}
\label{S3}

This section provides numerical results that illustrate the properties of our proposed significance test.
From a random vector of seven independent features $X=(X_1,\ldots,X_7) \sim \mathcal{N}(0, I_7)$, we consider the following data generating process: $Y = 3 + 4X_1 + X_1X_2 + 3X_3^2 + 2X_4X_5+ \epsilon$ where $\epsilon \sim N(0, \sigma^2)$ and $\sigma = 0.01$. Both variables $X_1$ and $X_3$ are first-order significant while $X_2$, $X_4$ and $X_5$ are second-order significant. Variables $X_6$ and $X_7$ have no influence on $Y$ and hence are non-significant. An intercept variable is added to the set of features. 
We generate a training, validation and testing sets of $100000$, $20000$ and $10000$ independent samples respectively.

A fully-connected neural network with two hidden layers and ReLU activation function is fitted to the training set using the Adam stochastic optimization method with step size 0.001 and exponential decay rates for the moment estimates of 0.9 and 0.999. We use a batch size of 32, a maximum number of 50 epochs and an early stopping criterion that stops the training when the validation error has not decreased by more than $10^{-2}$ for at least 5 epochs. The number of nodes in both hidden layers is chosen so as to minimize the validation loss. A network configuration of 150 nodes in the first layer and 50 in the second is found to perform best.

Given this choice of network architecture, we determine the optimal $\beta$ parameter using the model randomization procedure described in sub-section \ref{S22}. We search over values of $\beta$ from $10^{-6}$ to $10^{-1}$ and generate 20 random models per value of $\beta$. A value of $\beta = 10^{-2}$ is obtained.

We then apply first-order SFIT with $\alpha = 0.05$ and $\beta = 10^{-2}$. Only features $X_1$ and $X_3$ are returned as first-order significant as expected and the values of their test statistics and confidence intervals can be found in Table \ref{table:1}. In order to determine if it is necessary to look for second-order interactions, we test for the presence of any significant second-order effects as described in Algorithm \ref{algo 2} and find that there are. We then apply second-order SFIT with $\alpha = 0.05$ and $\beta = 10^{-2}$ and find that interactions $(1,2)$ and $(4,5)$ are significant. The values of the corresponding test statistics and confidence intervals can be found in Table \ref{table:1}. This means that features $X_2$, $X_4$ and $X_5$ are second-order significant. We finally test for the presence of any significant third-order effects and find that there is none meaning that we can stop the procedure. 

\begin{table}[h]
\small
\begin{center}
\caption{Test statistics $\hat{m}_j$ and their associated confidence intervals of the first and second-order significant features returned by the SFIT algorithm for $\alpha = 0.05$, $\beta = 10^{-2}$.}
\label{table:1}
\begin{tabularx}{\columnwidth}{ X|X|X } 
  Significant features or pairs  & $\hat{m}_j$ & 95\%-Confidence interval \\ 
 \hline
 $X_1$ & 1.13 & $[1.06,1.21]$ \\ 
 $X_3$ & 0.224 & $[0.185,0.265]$ \\ 
 $(X_1,X_2)$ & 0.01 & $[0.004,0.016]$ \\
 $(X_4,X_5)$ & 0.048 & $[0.035,0.060]$ \\
 \hline
\end{tabularx}
\end{center}
\end{table}

\begin{figure}[h]
\centering
\includegraphics[width=0.45\textwidth]{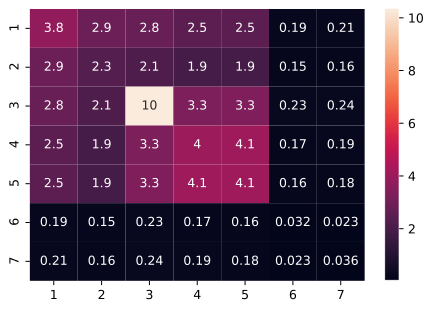}
\caption{Matrix $|W|^T|W|$ where $W$ is the matrix of parameters connecting the input layer to the first hidden layer of the fitted network.}
\label{fig1}
\end{figure}

As justified in section \ref{S23}, we suggest to look at the network parameters that connect the input layer to the first hidden layer in order to speed-up the search for significant interactions of features. A large entry at the position $(i,j)$ in this matrix indicates a likely significant interaction between features $i$ and $j$. As confirmed in Figure \ref{fig1}, the largest entry of the second row ($X_2$'s row) is indeed at the first column, pointing out interaction $(1,2)$ as the most significant one. The same is observed for the interaction between features $X_4$ and $X_5$ by looking at their corresponding rows. This confirms that information from the parameters of the network can indeed speed-up the search for interactions among features.

We now compare SFIT performance with the LOCO method. The output of LOCO tests can be found in Table \ref{table:2}. The first advantage of the SFIT method is its computation time. Indeed, it is an average 30 times faster to compute first-order SFIT over LOCO. Besides, even though the LOCO feature importance metric of non-informative features is smaller than the predictive ones, the LOCO method still fails to filter-out the two non-significant features. This confirms the importance of "regularizing" the test procedure and the introduction of the $\beta$ parameter that we propose to better control type-I errors.

\begin{table}[h]
\small
\begin{center}
\caption{Test statistics and their associated confidence intervals of the features selected by the LOCO algorithm for $\alpha = 0.05$.}
\label{table:2}
\begin{tabularx}{\columnwidth}{ X|X|X } 
 Significant features & Test statistic & 95\%-Confidence interval \\ 
 \hline
 $X_1$ & 2.51 & $[2.46, 2.57]$ \\ 
 $X_2$ & 0.434 & $[0.420, 0.448]$ \\ 
 $X_3$ & 2.18 & $[2.15, 2.20]$ \\ 
 $X_4$ & 0.732 & $[0.706, 0.754]$ \\ 
 $X_5$ & 0.740 & $[0.717, 0.767]$ \\ 
 $X_6$ & 0.023 & $[0.020, 0.025]$ \\ 
 $X_7$ & 0.006 & $[0.004, 0.008]$ \\ 
 \hline
\end{tabularx}
\end{center}
\end{table}

\begin{figure}[h]
\centering
\includegraphics[width=0.5\textwidth]{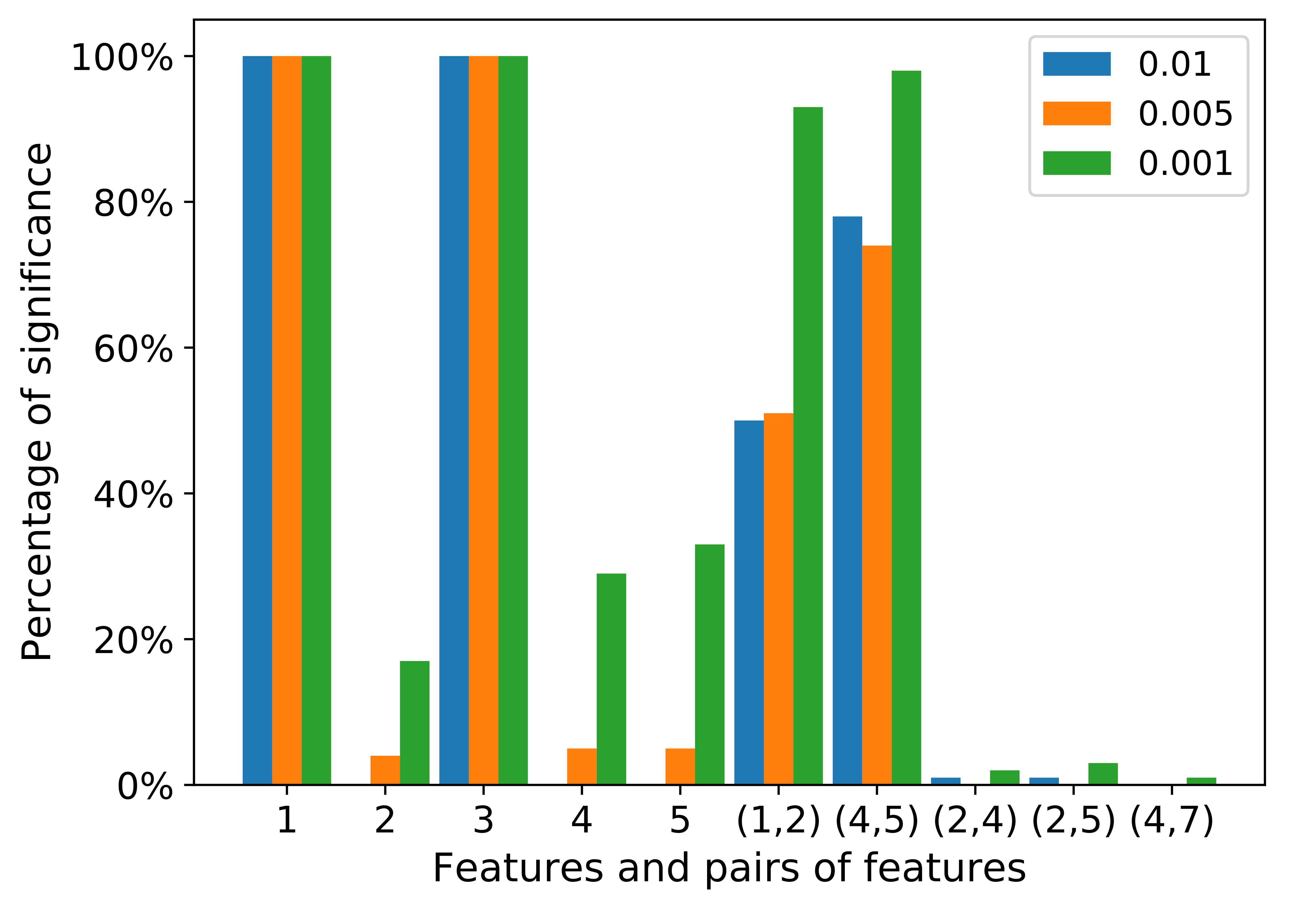}
\caption{Percentage of time the features or pairs of features are considered as significant over 100 tests for different values of $\beta$ and for $\alpha = 0.05$ and $n_2 = 10000$.}
\label{fig2.1}
\end{figure}

\begin{figure}[h]
\centering
\includegraphics[width=0.45\textwidth]{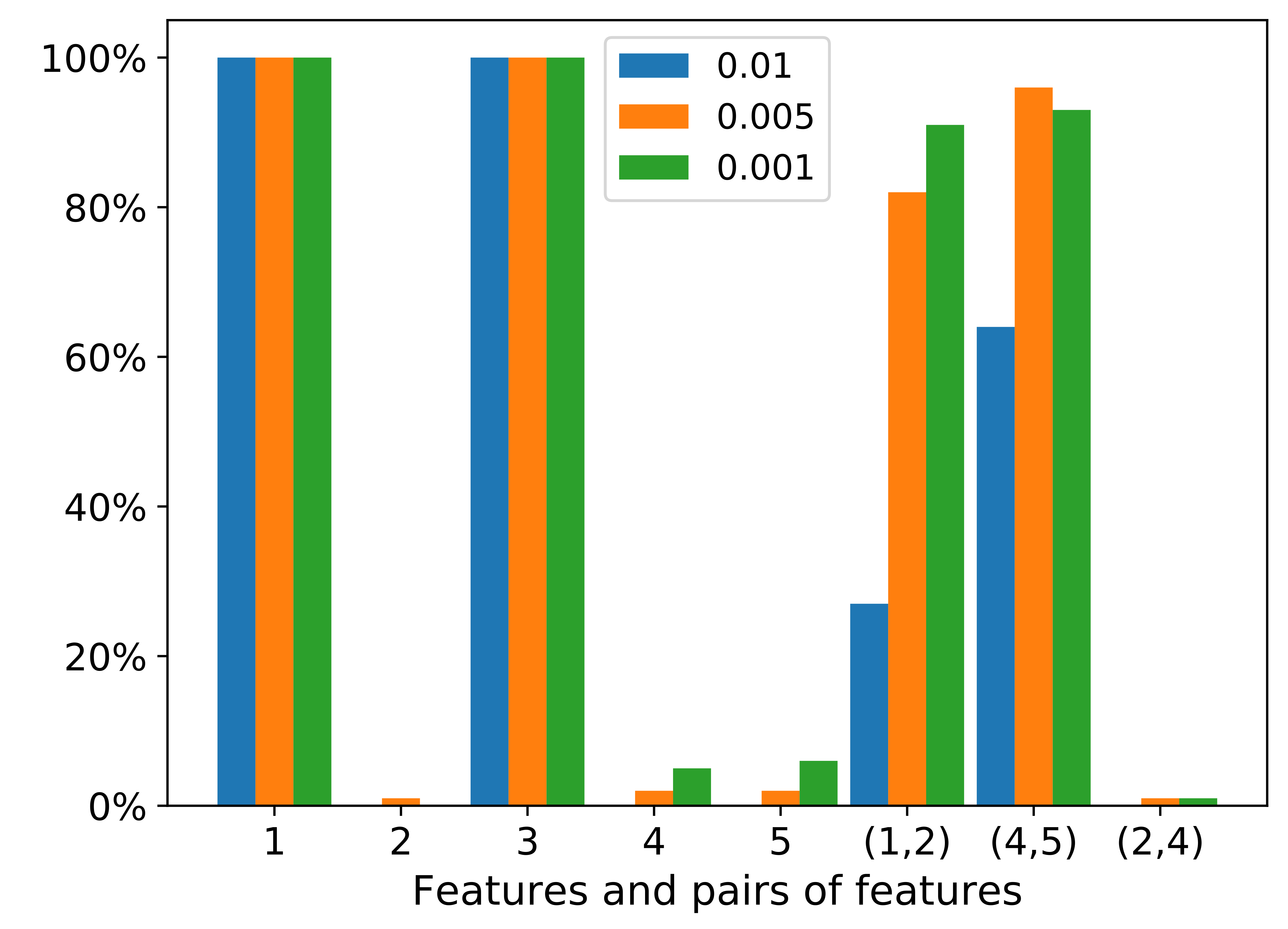}
\caption{Percentage of time the features or pairs of features are considered as significant over 100 tests for different values of $\beta$ and for $\alpha = 0.01$ and $n_2 = 10000$.}
\label{fig2.2}
\end{figure}

\begin{figure}[h]
\centering
\includegraphics[width=0.48\textwidth]{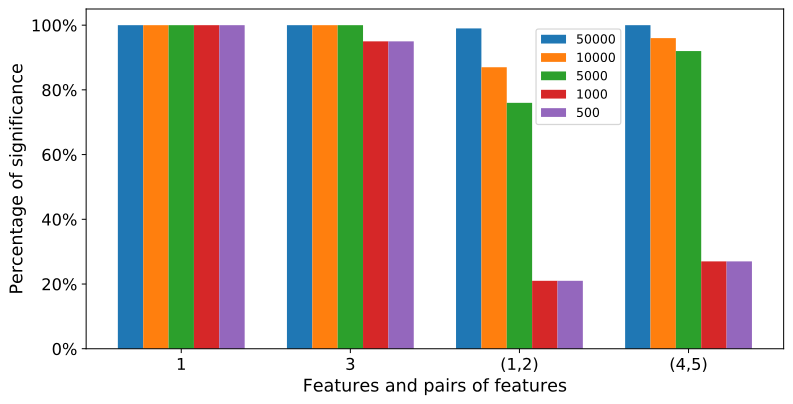}
\caption{Percentage of time the features or pairs of features are considered as significant over 100 tests for different values $n_2$ and for $\alpha = 0.05$, $\beta_1 = 0.01$ and $\beta_2 = 0.001$.}
\label{fig3}
\end{figure}

In order to better understand how the power and size of our proposed test procedure vary as a function of the parameters $\alpha$, $\beta$ and $n_2$, we repeat the testing procedure 100 times over 100 different training and testing sets. We record the fraction of time, each first order and second order features are considered as significant. In Figures \ref{fig2.1} and \ref{fig2.2}, we show how it varies with different choice of $\alpha$ and $\beta$. As we can see, $\beta$ seems to have a stronger effect on the power and size of the test than $\alpha$. The signals associated with first-order effects are strong so they can be well discriminated from noise using a value of $\beta$ as large as 0.01. However, because second-order effects have smaller signals, they are uncovered with a smaller value of $\beta \sim 0.001$. As a general guidance, we would recommend to use smaller values for the $\beta$ parameter when testing for higher-order effects. In Figure \ref{fig3}, we show the sensitivity of the test with respect to the test size $n_2$. The smaller the size of the inference set is, the smaller is the power of the corresponding test. We recommend to use a size of at least a couple thousands, larger number of test samples do not seem to improve further the accuracy of the test.



We illustrated through these simulations the accuracy of our test procedure. In the chosen regression setting, SFIT successfully discriminated the significant features and interactions from the non-informative ones. We highlighted its computational efficiency over the LOCO method and superior robustness to noise. Finally, we gave more insights into the effect of the method's parameters on its performance.

\section{Empirical results}
\label{S4}

In this section, the SFIT method is applied to the \href{https://archive.ics.uci.edu/ml/datasets/default+of+credit+card+clients}{UCI credit card defaults dataset} from \cite{yeh2009comparisons}. This dataset consists of 30,000 credit card holders payment data from a cash and credit card issuer in Taiwan in October 2005. The response variable is a binary default payment indicator (default = 1, no default = 0) and is associated with 23 explanatory variables. Description of the variables can be found in supplementary material.

As data pre-processing, we transform the categorical variables into dummies (this generates an overall number of 75 explanatory variables) and we normalize the continuous variables. The dataset is randomly splitted into a training (70\%), validation (20\%) and testing (10\%) set.

A fully-connected neural network with ReLU activation function is fitted to the training set by minimizing the binary cross-entropy loss using the Adam stochastic optimization method with step size 0.001 and exponential decay rates for the moment estimates of 0.9 and 0.999. We use a batch size of 32, a maximum number of 50 epochs and an early stopping criterion that stops the training when the validation error has not decreased by more than $10^{-3}$ for at least 10 epochs. 
The number of layers and nodes per hidden layers is chosen through grid-search so as to minimize the validation loss. This leads to a choice of architecture of 3 hidden layers with 100 nodes in the first layer, 50 in the second and 30 in the third. 

Because of the class imbalance (22.12\% of default observations), we weight the loss function during training to constraint the model to pay more attention to the under-represented class and we measure the model classification performance using the AUC score and the balanced classification accuracy.

We apply the SFIT method on the trained neural network model with $\alpha = 0.05$ and $\beta = 5e^{-2}$ found using the model randomization procedure described in section \ref{S22}. 24 variables (out of 75 explanatory variables) are returned as first-order significant and no second-order effect is found. The significant variables are displayed in Table \ref{table:4}, sorted by decreasing order of importance.

In order to assess the quality of the variable selection induced by the significance tests, we retrain the neural network on the subset of the 24 variables tested as significant by SFIT. We can observe no significant degradation of predictive accuracy of the model even with less than a third of the original variables as shown in Table \ref{table:5}. 

\begin{table}[h]
\small
\begin{center}
\caption{Test statistics of the significant SFIT variables for $\alpha = 0.05$ and $\beta = 5e^{-2}$}
\label{table:4}
\begin{tabular}{ c|c } 
 Significant features & Test statistic \\ 
 \hline
PAY\_0\_pay\_duly                         & 0.357 \\
PAY\_0\_payment\_delay\_for\_two\_month   & 0.353 \\
PAY\_0\_payment\_delay\_for\_three\_month & 0.300 \\
PAY\_2\_payment\_delay\_for\_three\_month & 0.235 \\
PAY\_5\_pay\_duly                         & 0.196 \\
PAY\_6\_payment\_delay\_for\_five\_month  & 0.136 \\
PAY\_0\_payment\_delay\_for\_four\_month  & 0.122 \\
PAY\_3\_pay\_duly                         & 0.119 \\
PAY\_2\_pay\_duly                         & 0.0966 \\
PAY\_6\_payment\_delay\_for\_two\_month   & 0.0946  \\
PAY\_5\_payment\_delay\_for\_seven\_month & 0.0924 \\
PAY\_6\_pay\_duly                         & 0.0794 \\
EDUCATION\_high\_school                   & 0.0665 \\
BILL\_AMT5                                & 0.0616 \\
PAY\_3\_payment\_delay\_for\_seven\_month & 0.0584 \\
PAY\_5\_payment\_delay\_for\_two\_month   & 0.0433 \\
PAY\_3\_payment\_delay\_for\_one\_month   & 0.0378 \\
BILL\_AMT4                                & 0.0331 \\
PAY\_6\_payment\_delay\_for\_three\_month & 0.0296 \\
PAY\_3\_payment\_delay\_for\_three\_month & 0.0290 \\
EDUCATION\_university                     & 0.0155 \\
PAY\_3\_payment\_delay\_for\_six\_month   & 0.0124 \\
PAY\_5\_payment\_delay\_for\_six\_month   & 0.00945 \\
PAY\_4\_payment\_delay\_for\_four\_month  & 0.00262 \\
 \hline
\end{tabular}
\end{center}
\end{table}

\begin{table}[h]
\small
\begin{center}
\caption{Predictive performance as measured by AUC and balanced accuracy of the neural network trained on the full set of 75 variables and on the subset of 24 variables selected by SFIT.}
\label{table:5}
\begin{tabularx}{\columnwidth}{ X|X|X }
 Metric & Neural network on all variables & Neural network on selected variables\\
 \hline
 AUC & 0.775 & 0.765\\ 
 Balanced accuracy & 0.707  & 0.701 \\ 
 \hline
\end{tabularx}
\end{center}
\end{table}

\FloatBarrier

\bibliography{biblio_SFIT}

\end{document}

%% file: math_commands.tex
\usepackage{amsmath,amsfonts,bm}

\def\eqref#1{equation~\ref{#1}}

\def\1{\bm{1}}

\DeclareMathAlphabet{\mathsfit}{\encodingdefault}{\sfdefault}{m}{sl}
\SetMathAlphabet{\mathsfit}{bold}{\encodingdefault}{\sfdefault}{bx}{n}













%% file: main.bbl
\begin{thebibliography}{}

\bibitem[Adebayo et~al., 2018]{adebayo2018sanity}
Adebayo, J., Gilmer, J., Muelly, M., Goodfellow, I., Hardt, M., and Kim, B.
  (2018).
\newblock Sanity checks for saliency maps.
\newblock In {\em Advances in Neural Information Processing Systems}, pages
  9505--9515.

\bibitem[Baehrens et~al., 2010]{baehrens2010explain}
Baehrens, D., Schroeter, T., Harmeling, S., Kawanabe, M., Hansen, K., and
  M{\~A}{\v{z}}ller, K.-R. (2010).
\newblock How to explain individual classification decisions.
\newblock {\em Journal of Machine Learning Research}, 11(Jun):1803--1831.

\bibitem[Barber et~al., 2015]{barber2015controlling}
Barber, R.~F., Cand{\`e}s, E.~J., et~al. (2015).
\newblock Controlling the false discovery rate via knockoffs.
\newblock {\em The Annals of Statistics}, 43(5):2055--2085.

\bibitem[Barber et~al., 2018]{barber2018robust}
Barber, R.~F., Cand{\`e}s, E.~J., and Samworth, R.~J. (2018).
\newblock Robust inference with knockoffs.
\newblock {\em arXiv preprint arXiv:1801.03896}.

\bibitem[Benjamini et~al., 2001]{benjamini2001control}
Benjamini, Y., Yekutieli, D., et~al. (2001).
\newblock The control of the false discovery rate in multiple testing under
  dependency.
\newblock {\em The annals of statistics}, 29(4):1165--1188.

\bibitem[Burns et~al., 2019]{burns2019interpreting}
Burns, C., Thomason, J., and Tansey, W. (2019).
\newblock Interpreting black box models with statistical guarantees.
\newblock {\em arXiv preprint arXiv:1904.00045}.

\bibitem[Candes et~al., 2018]{candes2018panning}
Candes, E., Fan, Y., Janson, L., and Lv, J. (2018).
\newblock Panning for gold:‘model-x’knockoffs for high dimensional
  controlled variable selection.
\newblock {\em Journal of the Royal Statistical Society: Series B (Statistical
  Methodology)}, 80(3):551--577.

\bibitem[Cortez and Embrechts, 2011]{cortez2011opening}
Cortez, P. and Embrechts, M.~J. (2011).
\newblock Opening black box data mining models using sensitivity analysis.
\newblock In {\em 2011 IEEE Symposium on Computational Intelligence and Data
  Mining (CIDM)}, pages 341--348. IEEE.

\bibitem[Datta et~al., 2016]{datta2016algorithmic}
Datta, A., Sen, S., and Zick, Y. (2016).
\newblock Algorithmic transparency via quantitative input influence: Theory and
  experiments with learning systems.
\newblock In {\em 2016 IEEE symposium on security and privacy (SP)}, pages
  598--617. IEEE.

\bibitem[Dixon and Mood, 1946]{dixon1946sign}
Dixon, W.~J. and Mood, A.~M. (1946).
\newblock The statistical sign test.
\newblock {\em Journal of the American Statistical Association},
  41(236):557--566.

\bibitem[Guidotti et~al., 2018]{guidotti2018survey}
Guidotti, R., Monreale, A., Ruggieri, S., Turini, F., Giannotti, F., and
  Pedreschi, D. (2018).
\newblock A survey of methods for explaining black box models.
\newblock {\em ACM computing surveys (CSUR)}, 51(5):93.

\bibitem[Henderson and Parmeter, 2015]{henderson2015applied}
Henderson, D.~J. and Parmeter, C.~F. (2015).
\newblock {\em Applied nonparametric econometrics}.
\newblock Cambridge University Press.

\bibitem[Horel and Giesecke, 2019]{horel2019towards}
Horel, E. and Giesecke, K. (2019).
\newblock Towards explainable ai: Significance tests for neural networks.
\newblock {\em arXiv preprint arXiv:1902.06021}.

\bibitem[Horel et~al., 2018]{horel2018sensitivity}
Horel, E., Mison, V., Xiong, T., Giesecke, K., and Mangu, L. (2018).
\newblock Sensitivity based neural networks explanations.
\newblock {\em arXiv preprint arXiv:1812.01029}.

\bibitem[Kononenko et~al., 2010]{kononenko2010efficient}
Kononenko, I. et~al. (2010).
\newblock An efficient explanation of individual classifications using game
  theory.
\newblock {\em Journal of Machine Learning Research}, 11(Jan):1--18.

\bibitem[Lei et~al., 2018]{Lei2018predictive}
Lei, J., G’Sell, M., Rinaldo, A., Tibshirani, R.~J., and Wasserman, L.
  (2018).
\newblock Distribution-free predictive inference for regression.
\newblock {\em Journal of the American Statistical Association},
  113(523):1094--1111.

\bibitem[Lundberg and Lee, 2017]{lundberg2017unified}
Lundberg, S.~M. and Lee, S.-I. (2017).
\newblock A unified approach to interpreting model predictions.
\newblock In {\em Advances in Neural Information Processing Systems}, pages
  4765--4774.

\bibitem[Olden and Jackson, 2002]{olden2002illuminating}
Olden, J.~D. and Jackson, D.~A. (2002).
\newblock Illuminating the “black box”: a randomization approach for
  understanding variable contributions in artificial neural networks.
\newblock {\em Ecological modelling}, 154(1-2):135--150.

\bibitem[Racine, 1997]{racine1997consistent}
Racine, J. (1997).
\newblock Consistent significance testing for nonparametric regression.
\newblock {\em Journal of Business \& Economic Statistics}, 15(3):369--378.

\bibitem[Ribeiro et~al., 2016]{ribeiro2016should}
Ribeiro, M.~T., Singh, S., and Guestrin, C. (2016).
\newblock Why should i trust you?: Explaining the predictions of any
  classifier.
\newblock In {\em Proceedings of the 22nd ACM SIGKDD international conference
  on knowledge discovery and data mining}, pages 1135--1144. ACM.

\bibitem[Shrikumar et~al., 2017]{shrikumar2017learning}
Shrikumar, A., Greenside, P., and Kundaje, A. (2017).
\newblock Learning important features through propagating activation
  differences.
\newblock In {\em Proceedings of the 34th International Conference on Machine
  Learning-Volume 70}, pages 3145--3153. JMLR. org.

\bibitem[Sundararajan et~al., 2017]{sundararajan2017axiomatic}
Sundararajan, M., Taly, A., and Yan, Q. (2017).
\newblock Axiomatic attribution for deep networks.
\newblock In {\em Proceedings of the 34th International Conference on Machine
  Learning-Volume 70}, pages 3319--3328. JMLR. org.

\bibitem[Tansey et~al., 2018]{tansey2018holdout}
Tansey, W., Veitch, V., Zhang, H., Rabadan, R., and Blei, D.~M. (2018).
\newblock The holdout randomization test: Principled and easy black box feature
  selection.
\newblock {\em arXiv preprint arXiv:1811.00645}.

\bibitem[Vuong, 1989]{vuong1989likelihood}
Vuong, Q.~H. (1989).
\newblock Likelihood ratio tests for model selection and non-nested hypotheses.
\newblock {\em Econometrica: Journal of the Econometric Society}, pages
  307--333.

\bibitem[Wei et~al., 2018]{wei2018margin}
Wei, C., Lee, J.~D., Liu, Q., and Ma, T. (2018).
\newblock On the margin theory of feedforward neural networks.
\newblock {\em arXiv preprint arXiv:1810.05369}.

\bibitem[Williamson et~al., 2017]{williamson2017nonparametric}
Williamson, B.~D., Gilbert, P.~B., Simon, N., and Carone, M. (2017).
\newblock Nonparametric variable importance assessment using machine learning
  techniques.
\newblock {\em UW Biostatistics Working Paper Series, Working Paper 422}.

\bibitem[Yeh and Lien, 2009]{yeh2009comparisons}
Yeh, I.-C. and Lien, C.-h. (2009).
\newblock The comparisons of data mining techniques for the predictive accuracy
  of probability of default of credit card clients.
\newblock {\em Expert Systems with Applications}, 36(2):2473--2480.

\bibitem[Zhang et~al., 2016]{zhang2016understanding}
Zhang, C., Bengio, S., Hardt, M., Recht, B., and Vinyals, O. (2016).
\newblock Understanding deep learning requires rethinking generalization.
\newblock {\em arXiv preprint arXiv:1611.03530}.

\end{thebibliography}
